%% file: main.tex
\documentclass{article} % For LaTeX2e
\usepackage{iclr2024_conference,times}

\input{math_commands.tex}

\usepackage[hidelinks]{hyperref}
\usepackage{subfiles}
\usepackage{url}
\usepackage{xcolor}
\usepackage{booktabs}       % professional-quality tables
\usepackage{amsfonts}       % blackboard math symbols
\usepackage{nicefrac}       % compact symbols for 1/2, etc.
\usepackage{microtype}      % microtypography
\usepackage{xcolor}         % colors
\usepackage{soul}
\usepackage{colortbl}
\usepackage{makecell}
\usepackage{listings}
\usepackage{graphicx}
\usepackage{multirow}
\usepackage{xspace}
\usepackage{tikz}
\usepackage{caption}
\usepackage{adjustbox}
\usepackage{rotating,tabularx}
\usepackage[noline,ruled,noend]{algorithm2e}

\usepackage{soul}
\usepackage{subcaption}
\usepackage{sidecap}
\def\*#1{\mathbf{#1}}

\usepackage{comment}
\usepackage{amsthm}
\theoremstyle{plain}
\newtheorem{theorem}{Theorem}[section]
\newtheorem{proposition}[theorem]{Proposition}

\newtheorem{manualtheoreminner}{Proposition}
\newenvironment{manualtheorem}[1]{%
  \renewcommand\themanualtheoreminner{#1}%
  \manualtheoreminner
}{\endmanualtheoreminner}

\title{Scaling for Training Time and Post-hoc Out-of-distribution Detection Enhancement}

\author{Kai Xu$^{1}$, Rongyu Chen$^1$, Gianni Franchi$^2$, Angela Yao$^{1}$ \\
$^1$National University of Singapore\\
$^2$U2IS, ENSTA Paris, Institut polytechnique de Paris\\
{\scriptsize \texttt{\{kxu,rchen,ayao\}@comp.nus.edu.sg} \quad \texttt{gianni.franchi@ensta-paris.fr}} \\
}

\makeatletter
\DeclareRobustCommand\onedot{\futurelet\@let@token\@onedot}
\def\@onedot{\ifx\@let@token.\else.\null\fi\xspace}

\def\ie{\emph{i.e}\onedot}

\makeatother
\definecolor{Maroon}{cmyk}{0, 0.87, 0.68, 0.32}

\newcommand{\logit}{\vz}  %logit value
\newcommand{\pF}{\va}     %penultimate activation

\begin{document}

\maketitle

\begin{abstract}
The capacity of a modern deep learning system to determine if a sample falls within its realm of knowledge is fundamental and important.
In this paper, we offer insights and analyses of recent state-of-the-art out-of-distribution (OOD) detection methods - extremely simple activation shaping (ASH). We demonstrate that activation pruning has a detrimental effect on OOD detection, while activation scaling enhances it.
Moreover, we propose SCALE, a simple yet effective post-hoc network enhancement method for OOD detection, which attains state-of-the-art OOD detection performance without compromising in-distribution (ID) accuracy. By integrating scaling concepts into the training process to capture a sample's ID characteristics, we propose \textbf{I}ntermediate Tensor \textbf{SH}aping (ISH), a lightweight method for training time OOD detection enhancement. We achieve AUROC scores of +1.85\% for near-OOD and +0.74\% for far-OOD datasets on the OpenOOD v1.5 ImageNet-1K benchmark. Our code and models are available at \url{https://github.com/kai422/SCALE}.

\end{abstract}

\section{Introduction}
In deep neural networks, out-of-distribution (OOD) detection distinguishes samples which deviate from the training distribution. Standard OOD detection concerns semantic shifts~\citep{DBLP:conf/nips/YangWZZDPWCLSDZ22/OpenOOD, DBLP:journals/corr/abs-2306-09301/OpenOODv1.5}, where OOD data is defined as test samples from semantic categories unseen during training.  Ideally, the neural network should be able to reject such samples as being OOD, while still maintaining strong performance on in-distribution (ID) test samples belonging to seen training categories.

Methods for detecting OOD samples work by scoring network outputs such as logits or softmax values~\citep{DBLP:conf/iclr/HendrycksG17/MSP,DBLP:conf/icml/HendrycksBMZKMS22/MLS}, post-hoc network adjustment during inference to improve OOD scoring~\citep{DBLP:conf/eccv/SunL22a/DICE, DBLP:conf/nips/SunGL21/ReAct,DBLP:conf/iclr/DjurisicBAL23/ASH}, or by adjusting model training~\citep{DBLP:conf/icml/WeiXCF0L22/LogitNorm,DBLP:conf/iclr/MingSD023/CIDER,DBLP:journals/corr/abs-1802-04865/ConfBranch}. These approaches can be used either independently or in conjunction with one another.  Typically, post-hoc adjustments together with OOD scoring is the preferred combination since it is highly effective at discerning OOD samples with minimal ID drop and can also be applied directly to already-trained models off-the-shelf.  Examples include ReAct~\citep{DBLP:conf/nips/SunGL21/ReAct}, DICE~\citep{DBLP:conf/eccv/SunL22a/DICE} and more recently, ASH~\citep{DBLP:conf/iclr/DjurisicBAL23/ASH}.

On the surface, each method takes different and sometimes even contradictory approaches.  
ReAct rectifies penultimate activations which exceed a threshold; ASH, on the other hand, prunes penultimate activations that are too low while amplifying remaining activations.
While ASH currently achieves state-of-the-art performance, it lacks a robust explanation of its underlying operational principles. This limitation highlights the need for a comprehensive explanatory framework.

This work seeks to understand the working principles behind ASH. Through observations and mathematical derivations, we reveal that OOD datasets tend to exhibit a lower rate of pruning due to distinct mean and variance characteristics. We also demonstrate the significant role of scaling in enhancing OOD detection in ASH, while highlighting that the lower-part pruning approach, in contrast to ReAct, hinders the OOD detection process.
This understanding leads to new state-of-the-art results by leveraging scaling, achieving significant improvements without compromising on ID accuracy. 

Through the lens of studying the distributions, we highlight the importance of scaling as a key metric for assessing a sample's ID nature. We integrate this concept into the training process, hypothesizing the feasibility of shaping the ID-ness objective even without the inclusion of OOD samples.
The ID-ness objective introduces an optimization weighting factor for different samples through proposed intermediate tensor shaping (ISH). Remarkably, ISH achieves outstanding performance in both near-OOD and far-OOD detection tasks, with only one-third of the training effort required compared to current state-of-the-art approaches.

Our contributions can be summarized as follows:
\begin{itemize}
    \item We analyze and explain the working principles of pruning and scaling for OOD detection and reveal that pruning, in some scenario, actually hurts OOD detection. 
    \item Based on our analysis, we devise SCALE, a new post-hoc network enhancement method for OOD detection, which achieves state-of-the-art results on OOD detection without any ID accuracy trade-off.
    \item By incorporating scaling concepts into the training process to capture a sample's ID characteristics, we introduce ISH, a lightweight and innovative method for improving OOD detection during training. ISH yields remarkable OOD detection results.
\end{itemize}

\begin{figure}
    \centering 
	\includegraphics[width=0.8\textwidth]{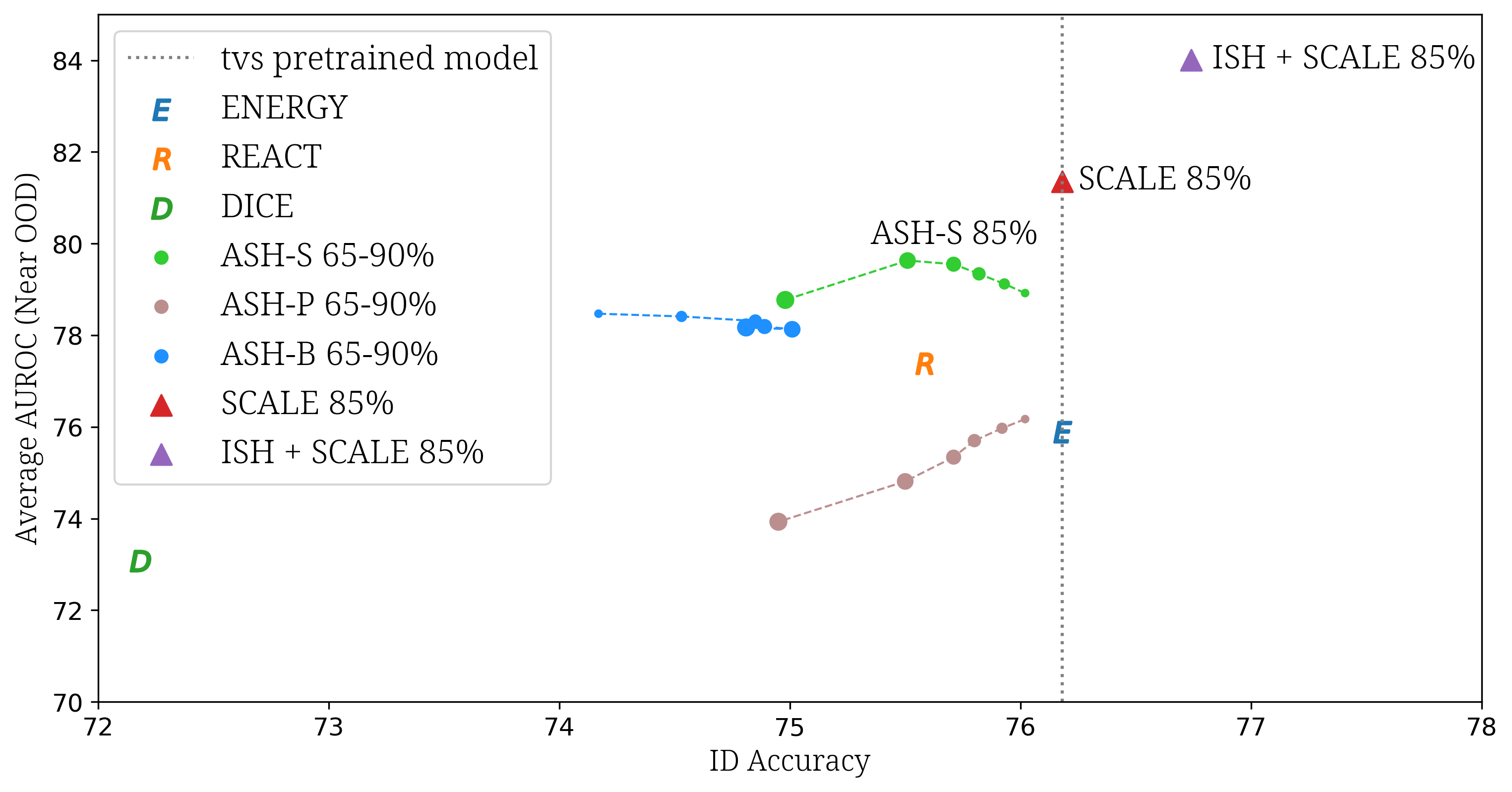}
	\caption{\textbf{ID-OOD Trade-off on ImageNet on Near-OOD Dataset.} Unlike existing methods such as ASH, ReAct and Dice, our proposed SCALE does not have any ID accuracy trade-off while improving OOD accuracy. Our training methods, ISH, achieves outstanding OOD results by emphasizing the training of samples with high ID characteristics. 
 }\label{fig:id-ood-tradeoff}
\end{figure}

\section{Related Work}

\textbf{OOD scoring methods} indicate how likely a sample comes from the training distribution, \ie is in-distribution, based on sample features or model outputs.  From a feature perspective, ~\cite{DBLP:conf/nips/LeeLLS18/MDS} proposed to score a sample via the minimum Mahalanobis distance of that sample's features to the nearest ID class centroid. For model outputs, two common variants are based on the maximum softmax prediction~\citep{DBLP:conf/iclr/HendrycksG17/MSP} and the maximum logit scores~\citep{DBLP:conf/icml/HendrycksBMZKMS22/MLS}. The raw softmax or logit scores are susceptible to the overconfidence issue, therefore,~\cite{DBLP:conf/nips/LiuWOL20/EBO} proposed to use an energy-based function to transform the logits as an improved score.  A key benefit of deriving OOD scores from feature or model outputs is that it does not impact the model or the inference procedure, so the ID accuracy will not be affected.

\textbf{Post-hoc model enhancement methods} modify the inference procedure to improve OOD detection and are often used together with OOD scoring methods.  
Examples include ReAct~\citep{DBLP:conf/nips/SunGL21/ReAct}, which rectifies the penultimate activations for inference,  DICE~\citep{DBLP:conf/eccv/SunL22a/DICE}, which sparsifies the network's weights in the last layer, and 
ASH~\citep{DBLP:conf/iclr/DjurisicBAL23/ASH}, which scales and prunes the penultimate activations. Each of these methods is then combined with energy-based score~\citep{DBLP:conf/nips/LiuWOL20/EBO} to detect the OOD data.  While effective at identifying OOD data, these methods have a reduced ID accuracy as the inference procedure is altered.  
Our proposed SCALE is also post-hoc model enhancement, while our ID accuracy will not be affected, where we applies different scaling factor based on sample's activations shape, which do not alter the ID estimates for single sample, but emphasize difference among samples.

\paragraph{Training-time model enhancement} techniques aims to make OOD data more distinguishable directly at training.  Various strategies including the incorporation of additional network branches~\citep{DBLP:journals/corr/abs-1802-04865/ConfBranch}, alternative training strategies~\citep{DBLP:conf/icml/WeiXCF0L22/LogitNorm}, or data augmentation~\citep{DBLP:conf/nips/PintoYLTD22/RegMixup,DBLP:conf/iclr/HendrycksMCZGL20/AugMix}.  The underlying assumption behind each of these techniques is training towards OOD detection objective can provide more discriminative features for OOD detection.
A significant drawback of training-time enhancement is the additional computational cost. For example, AugMix~\citep{DBLP:conf/iclr/HendrycksMCZGL20/AugMix} requires double training time and extra GPU memory cost. 
Our intermediate tensor shaping (ISH) improves the OOD detection with one-third of the computational cost compares to the most lightweight method, without modifying model architecture.

\textbf{Intermediate tensor shaping:}
Activation shaping have been explored in deep learning for various purposes. DropOut is the first to utilize this idea by sparsifying the activations for regularization. Similar ideas has been applied on~\cite{DBLP:conf/iclr/LiYBLRRYCYGK23/Lazy} for transformers. Activation shaping can also help efficient training and inference by compression~\citep{DBLP:conf/icml/KurtzKGMCGLMSA20/Inducing, DBLP:conf/cvpr/ChenLTYZH23/SparseViT}.
Shaping operations on intermediate tensors differ from those on activations. Activation shaping affects both forward pass inference and backward gradient computation during training. In contrast, shaping intermediate tensors exclusively influences the backward gradient computation. Since intermediate tensors tend to consume a significant portion of GPU memory, techniques for compressing intermediate tensors have gained widespread use in memory-efficient training, all without altering the forward pass.~\citep{DBLP:conf/nips/EvansA21/AC-GC, DBLP:conf/icml/LiuZWCCHCLTGMC22/GACT,  DBLP:conf/iclr/ChenXWCY23/DropIT}. 

\section{Activation Scaling for Post-hoc Model Enhancement}

We start by presenting the preliminaries of Out-of-Distribution (OOD) detection in Sec. 3.1 to set the stage for our subsequent discussion and analysis of the ASH method in Sec. 3.2. The results of our analysis directly leads to our own OOD criterion in Sec. 3.3.  Finally, we introduce our intermediate tensor shaping approach for training time OOD detection enhancement in Sec. 3.4.

\subsection{Preliminaries}

While OOD is relevant for many domains, we follow previous works~\citep{DBLP:conf/nips/YangWZZDPWCLSDZ22/OpenOOD} and focus specifically on semantic shifts in image classification.
During training, the classification model is trained with ID data that fall into a pre-defined set of $K$ semantic categories: $\forall (\vx,y)\sim \mathcal{D}_{\text{ID}}, y \in \mathcal{Y}_{\text{ID}}$. During inference, there are both ID and OOD samples; the latter are samples drawn from categories unobserved during training, 
\ie $\forall (\vx,y)\sim \mathcal{D}_{\text{OOD}}, y \notin \mathcal{Y}_{\text{ID}}$.

Now consider a neural network consisting of two parts: a feature extractor $f(\cdot)$, and a linear classifier parameterized by weight matrix $\mathbf{W}\in \mathbb{R}^{K\times D}$ and a bias vector $\vb \in \mathbb{R}^{D}$. The network logit can be mathematically represented as

\begin{equation}
    \logit = \mathbf{W} \cdot \pF + \vb, \qquad \pF = f(\vx), 
\end{equation}

where $\pF\in\mathbb{R}^{D}$ is the $D$-dimensional feature vector in the penultimate layer of the network and $\logit \in \mathbb{R}^{K}$ is the logit vector from which the class label can be estimated by $\hat{y} = \argmax(\logit)$.  In line with other OOD literature~\citep{DBLP:conf/nips/SunGL21/ReAct}, an individual dimension of feature $\pF$, denoted with index $j$ as $\pF_j$, is referred to as an ``activation''. 

For a given test sample $\vx$, an OOD score can be calculated to indicate the confidence that $\vx$ is in-distribution. By convention, scores above a threshold $\tau$ are ID, while those equal or below are considered OOD. A common setting is the energy-based OOD score $S_{\textit{EBO}}(\vx)$ together with indicator function $G(\cdot)$ that applies the thresholding~\citep{DBLP:conf/nips/LiuWOL20/EBO}:

\begin{align}\label{eq:OODscore}
G(\vx; \tau) =
\begin{cases}
0 & \quad \text{if } S_{\textit{EBO}}(\vx) \leq \tau \quad(\textit{OOD}), \\
1 & \quad \text{if } S_{\textit{EBO}}(\vx) > \tau \quad (\textit{ID}),
\end{cases},  \qquad  S_{\textit{EBO}}(\vx) = T\cdot \text{log}\sum_k^K e^{\logit_k/T},
\end{align}

where $T$ is a temperature parameter, $k$ is the logit index for the $K$ classes.

\subsection{Analysis on ASH:}\label{sec:analysis}
A state-of-the-art method for OOD detection is ASH~\citep{DBLP:conf/iclr/DjurisicBAL23/ASH}. ASH stands for activation shaping and is a simple post-hoc method that applies a rectified scaling to the feature vector $\pF$. Activations in $\pF$ up to the $p^{\text{th}}$ percentile across the $D$ dimensions are rectified (``pruned'' in the original text); activations above the $p^{\text{th}}$ percentile are scaled.  More formally, ASH introduces a shaping function $s_f$ that is applied to each activation $\pF_j$ in a given sample.  If we define $P_{p}(\pF)$ as the $p^{\text{th}}$ percentile of the elements in $\pF$, ASH produces the logit $\logit_{\text{ASH}}$:   

\begin{equation}\label{eq:ASHlogit}
    \logit_{\text{ASH}} = \mathbf{W} \cdot \left(\pF \circ s_f(\pF)\right) + \vb, \quad \text{where } s_f (\pF)_j = \begin{cases}
0 & \quad \text{if } \pF_j \leq P_{p}(\pF), \\
\exp(r) & \quad \text{if } \pF_j > P_{p}(\pF),
\end{cases}, \quad 
\end{equation}

and $\circ$ denotes an element-wise matrix multiplication, and the scaling factor $r$ is defined as the ratio of the sum of all activations versus the sum of un-pruned activations in $\pF$:

\begin{equation}\label{eq:scalefactor}
    r = \frac{Q}{Q_p}, \qquad \text{where } Q = {\sum_{j}^{D}{\va_j}} \qquad \text{ and }\; Q_p = \!\!\!\! \sum_{\va_j>P_{p}(\pF)}{\va_j}.
\end{equation}

Since $Q_p\leq Q$, the factor $r \geq 1$; the higher the percentile $p$, \ie the greater the extent of pruning, the smaller $Q_p$ is with respect to $Q$ and the larger the scaling factor $r$.  To distinguish OOD data, ASH then passes the logit from Eq.~\ref{eq:ASHlogit} to score and indicator function as given in Eq.~\ref{eq:OODscore}.

While ASH is highly effective, the original paper has no explanation of the working mechanism\footnote{In fact, the authors put forth a call for explanation in their Appendix L.}. We analyze the rectification and scaling components of ASH below and reveal that scaling helps to separate ID versus OOD energy scores, while rectification has an adverse effect. 

\vspace{2em}

\begin{minipage}[t!]{0.30\textwidth}
\vspace{-5em}
\centering
\resizebox{0.8\textwidth}{!}{%
\begin{tabular}{l|c}
\toprule
Dataset & $p$ value \\ \midrule
ImageNet & 0.296 \\
SSB-hard & 0.262 \\
NINCO & 0.181 \\ 
iNaturalist & 0.083 \\
Textures & 0.099 \\
OpenImage-O & 0.155 \\
\bottomrule
  \end{tabular}
  }
  \captionof{table}{Average $p$-values for all samples under Chi-square test; values greater than 0.05 verifies that a Gaussian assumption is reasonable. }\label{tab:assumption_guassian}
\end{minipage}
\hfill
\begin{minipage}[t]{0.33\textwidth}

\includegraphics[width=\linewidth]{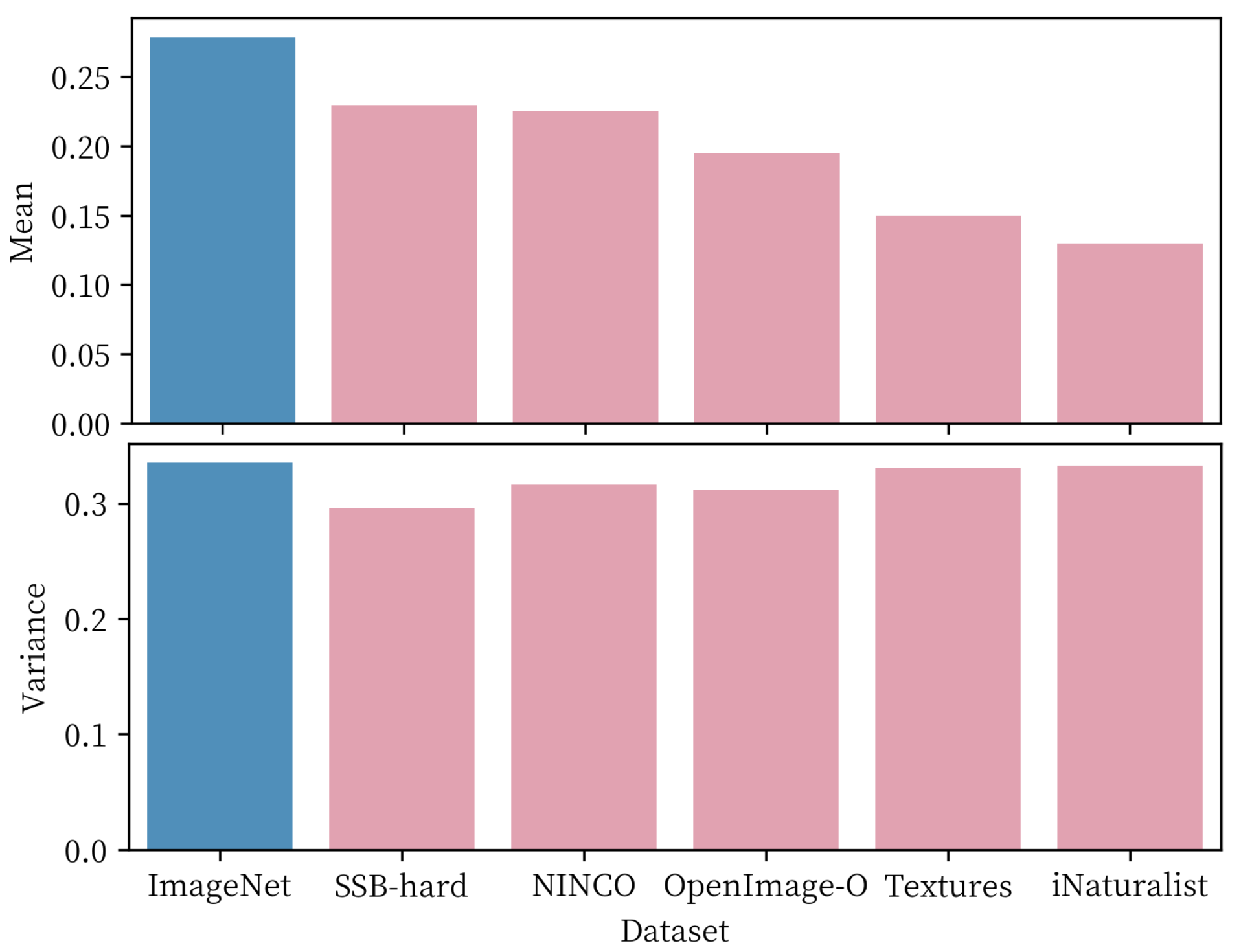}
\captionof{figure}{Mean and Variance of pre-ReLU activations for ID (blue) vs. OOD datasets (pink).}\label{fig:assumption_m_v}
  
\end{minipage}
\hfill
\begin{minipage}[t]{0.33\textwidth}

    \includegraphics[width=\linewidth]{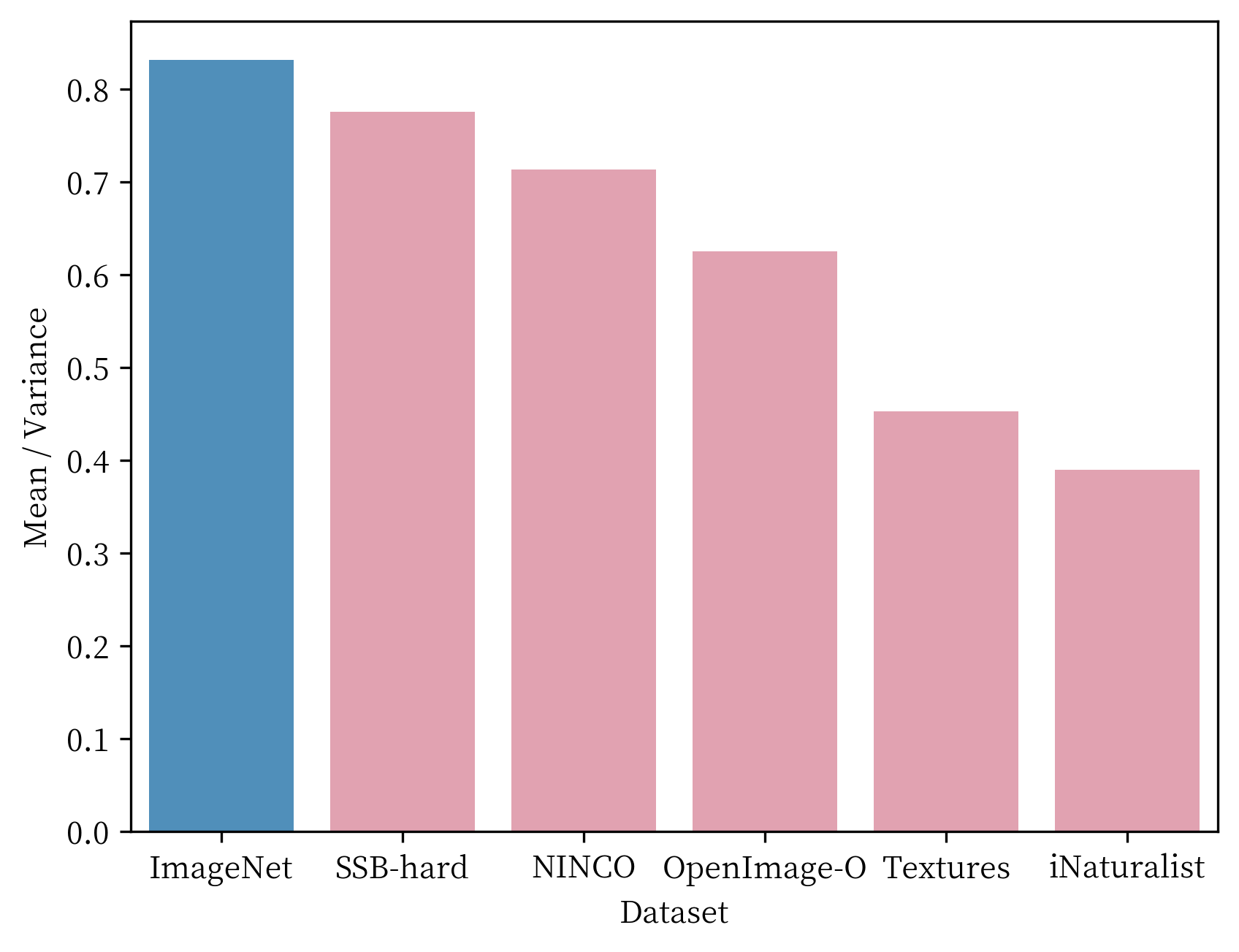}
    \captionof{figure}{$\mu/\sigma$ of pre-ReLU activations 
    for ID (blue) vs. OOD (pink).
}\label{fig:assumption_mdv}

\end{minipage}

\vspace{1em}
\textbf{Assumptions:} Our analysis is based on two assumptions.  (1) The penultimate activations of ID and OOD samples follow two differing rectified Gaussian distributions parameterized by $(\mu^{\text{ID}}, \sigma^{\text{ID}})$ and  $(\mu^{\text{OOD}}, \sigma^{\text{OOD}})$. The Gaussian assumption is commonly used in the literature~\citep{DBLP:conf/nips/SunGL21/ReAct}and we verify it in Tab. \ref{tab:assumption_guassian}; the rectification follows naturally if a ReLU is applied as the final operation of the penultimate layer.  (2) Normalized ID activations are higher than that of OOD activations; this assumption is supported by \citep{DBLP:conf/nips/LiuWOL20/EBO}
, who suggested that well-trained networks have higher responses to samples resembling those seen in training.
Fig.~\ref{fig:assumption_m_v} and Fig.~\ref{fig:assumption_mdv} visualize statistical corroboration of these assumptions. 

\begin{proposition}
\label{prop_ratio}
Assume that ID activations $\pF_j^{(\textit{ID})} \sim \mathcal{N}^{R}(\mu^{\textit{ID}}, \sigma^{\textit{ID}})$ and OOD activations $\pF_j^{(\textit{OOD})} \sim \mathcal{N}^{R}(\mu^{\textit{OOD}}, \sigma^{\textit{OOD}})$ where $\mathcal{N}^{R}$ denotes a rectified Gaussian distribution. 
If $\mu^\text{ID}/\sigma^\text{ID}>\mu^\text{OOD}/\sigma^\text{OOD}$, then there is a range of percentiles $p$
for which a factor $ C(p)=\frac{\varphi(\sqrt{2}\operatorname {erf} ^{-1}(2p-1))}{1-\Phi(\sqrt{2}\operatorname {erf} ^{-1}(2p-1))}$ is large enough such that ${Q_p^{\text{ID}}}/{Q^{\text{ID}}}<{Q_p^{\text{OOD}}}/{Q^{\text{OOD}}}$. 
\end{proposition}

\begin{figure}[t!]
  \centering

  \begin{subfigure}{0.31\textwidth}
    \includegraphics[width=\linewidth]{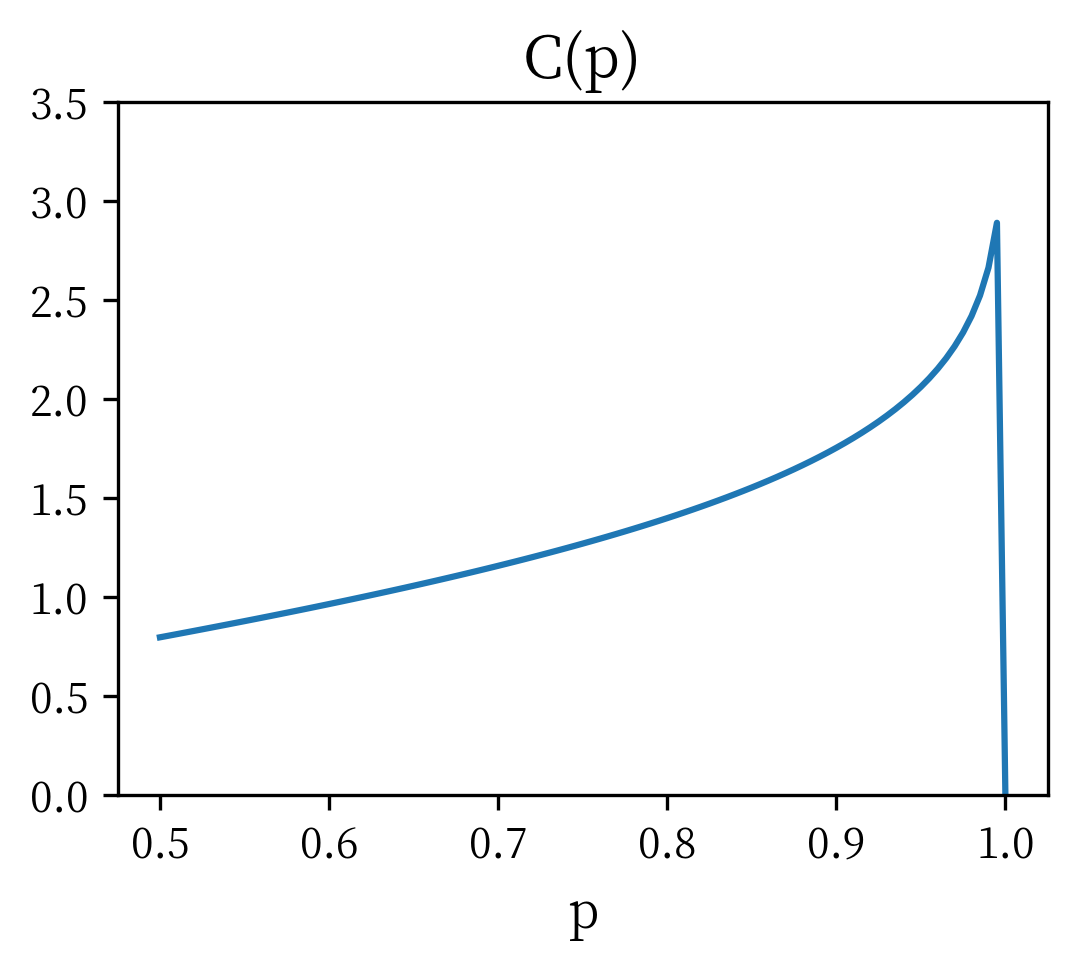}
  \end{subfigure}\hfil
  \begin{subfigure}{0.33\textwidth}
  
    \includegraphics[width=\linewidth]{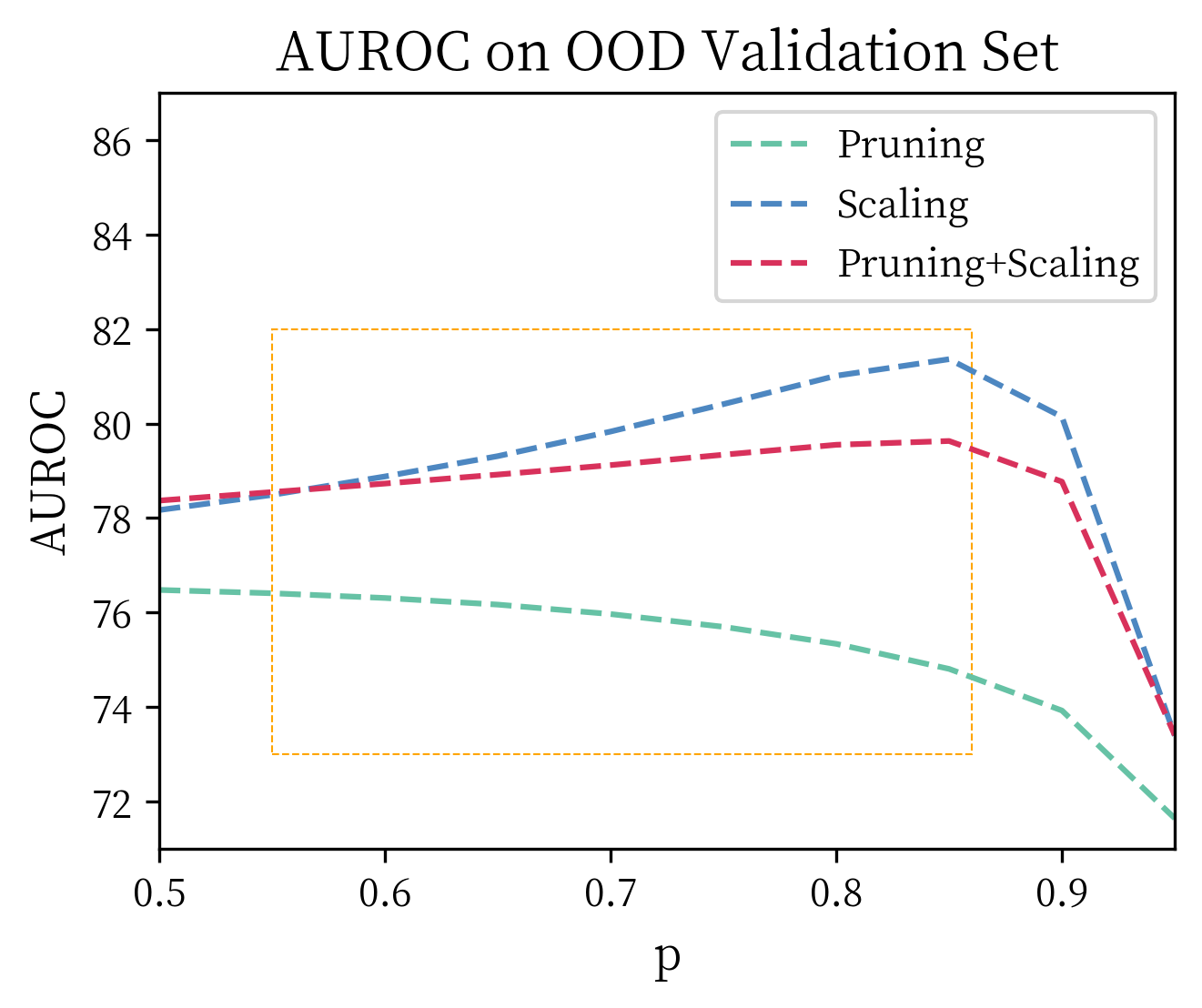}
  \end{subfigure}\hfil
  \begin{subfigure}{0.33\textwidth}
    \vspace{1em}
    \includegraphics[width=\linewidth]{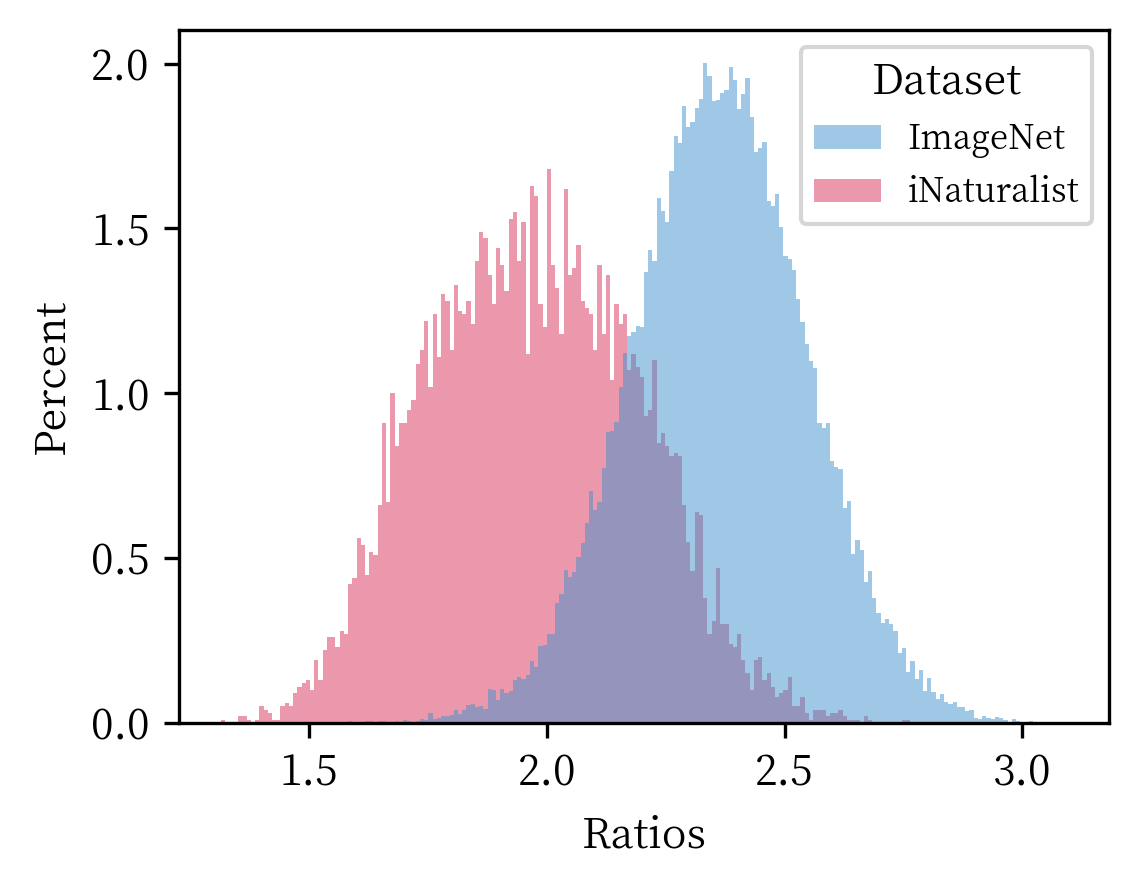}
  \end{subfigure}
  
  \caption{(a) 
  The relationship between the parameter $C(p)$ and the percentile $p$. A higher value of $C(p)$ indicates better separation of scales. (b) AUROC vs. percentile $p$. Up to $p=0.85$, as highlighted by orange box, AUROC for scaling increases while for pruning it decreases. The results of ASH sit between the two as the method is a combination of pruning plus scaling. (c) Histograms of scales $Q/Q_p$ for ID dataset (ImageNet) and  OOD dataset (iNaturalist) exhibit a clear separation from each other. }
  \label{fig:percentile}
\end{figure}

The full proof is given in Appendix~\ref{app:proof}. Above, $\varphi$ and $\Phi$ denote the probability density function and cumulative distribution function of the standard normal distribution, respectively.  The factor $C(p)$, plotted in Fig.~\ref{fig:percentile}a, relates the percentile of activations that distinguishes ID from OOD data.

\textbf{Rectification (Pruning)} 
The relative reduction of activations can be expressed as:

\begin{equation}
    D^\textit{Pruning} = (Q - Q_p)/{Q}.
\end{equation}

Note that a reduction in activations also leads to a reduction in the OOD energy. 
Since ${Q_p^{\textit{ID}}}/{Q^{\textit{ID}}}<{Q_p^{\textit{OOD}}}/{Q^{\textit{OOD}}}$, it directly implies that the decrease in ID samples will be greater than that in OOD samples, denoted as $D^\textit{Pruning}_{\textit{ID}}>D^\textit{Pruning}_{\textit{OOD}}$.
From this result, we can show that the expected value of the relative decrease in energy scores with rectification will be greater for ID samples than OOD samples following the Remark 2 in \cite{DBLP:conf/nips/SunGL21/ReAct}, which illustrates that the changes in logits is 
proportional to the changes in activations.  

Our result above shows that rectification or pruning creates a greater overlap in energy scores between ID and OOD samples, making it more difficult to distinguish them.  
Empirically, this result is shown in Fig.~\ref{fig:percentile}b, where AUROC steadily decreases with stand-alone pruning as the percentile $p$ increase.

\textbf{Scaling} on the other hand behaves in a manner opposite to the derivation above and enlarges the separation between ID and OOD scores.

Given ${Q_p^{\textit{ID}}}/{Q^{\textit{ID}}}<{Q_p^{\textit{OOD}}}/{Q^{\textit{OOD}}}$ and $r = Q/Q_q$, we have $r^\textit{ID} > r^\textit{OOD}$, which motivates the separation on $r$ between ID and OOD, Fig. \ref{fig:percentile}c depicts the histograms for these respective distributions, they are well separated and therefore scale activations of ID and OOD samples differently. The relative increase on activation can be expressed as:
\begin{equation}
    I^\text{Scaling} = (r - 1)
\end{equation}

where we can get $I^\text{scaling}_\text{ID} > I^\text{scaling}_\text{OOD}$. This increase is then transferred to logit spaces $\logit$ and energy-based scores $S_\textit{EBO(ID)}$ and $S_\textit{EBO(OOD)}$, which increase the gap between ID and OOD samples.

\textbf{Discussion on percentile $p$:} \ Note that $C(p)$ does not monotonically increasing with respect to $p$ (see Fig.~\ref{fig:percentile}a).  When $p \approx 0.95$, there is an inflection point and $C(p)$ decreases. A similar inflection follows on the AUROC for scaling (see Fig.~\ref{fig:percentile}b), though it is not exactly aligned to $C(p)$.  The difference is likely due to the approximations made to estimate $C(p)$.  Also, as $p$ gets progressively larger, fewer activations ($D=2048$ total activations) are considered for estimating $r$, leading to unreliable logits for the energy score.  Curiously, pruning also drops off, which we believe to come similarly from the extreme reduction in activations.

\subsection{SCALE Criterion for OOD Detection}\label{sec:scale}

From our analyses and findings above, we propose a new post-hoc model enhancement criterion, which we call \textit{SCALE}. As the name suggests, it shapes the activation with (only) a scaling:    
\begin{equation}
    \logit' = \mathbf{W} \cdot \left(\pF \circ s_f(\pF)\right) + \vb, \qquad \text{where } s_f (\pF)_j = \exp(r)\ \text{ and }\ r = \frac{\sum_{j}{\va_j}}{\sum_{\va_j>P_{p}(\pF)}{\va_j}}.
\end{equation}

\begin{figure}[tp]%
    \centering
    \begin{subfigure}[t]{0.50\linewidth}
        \centering
        \includegraphics[width=\linewidth]{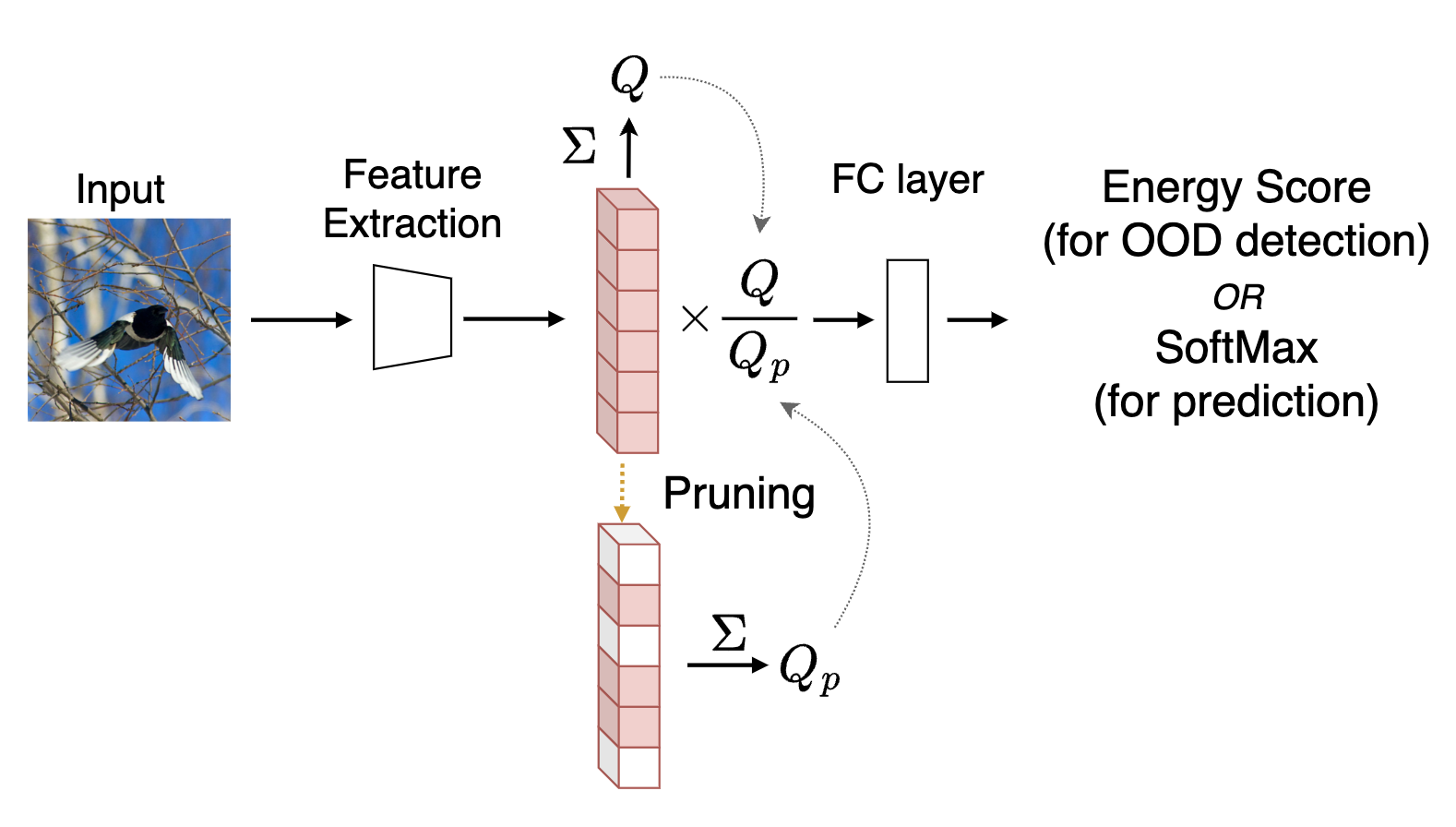}
        \caption{Demonstration of SCALE post-hoc model improvement. We prune activations to calculate the scaling factor. The original activations are then multiplied by the computed scales before fed into the fully connected layer. }\label{fig:scale}
    \end{subfigure}\hfil% equal to outside spacing
    \begin{subfigure}[t]{0.45\linewidth}
        \centering
        \includegraphics[width=\linewidth]{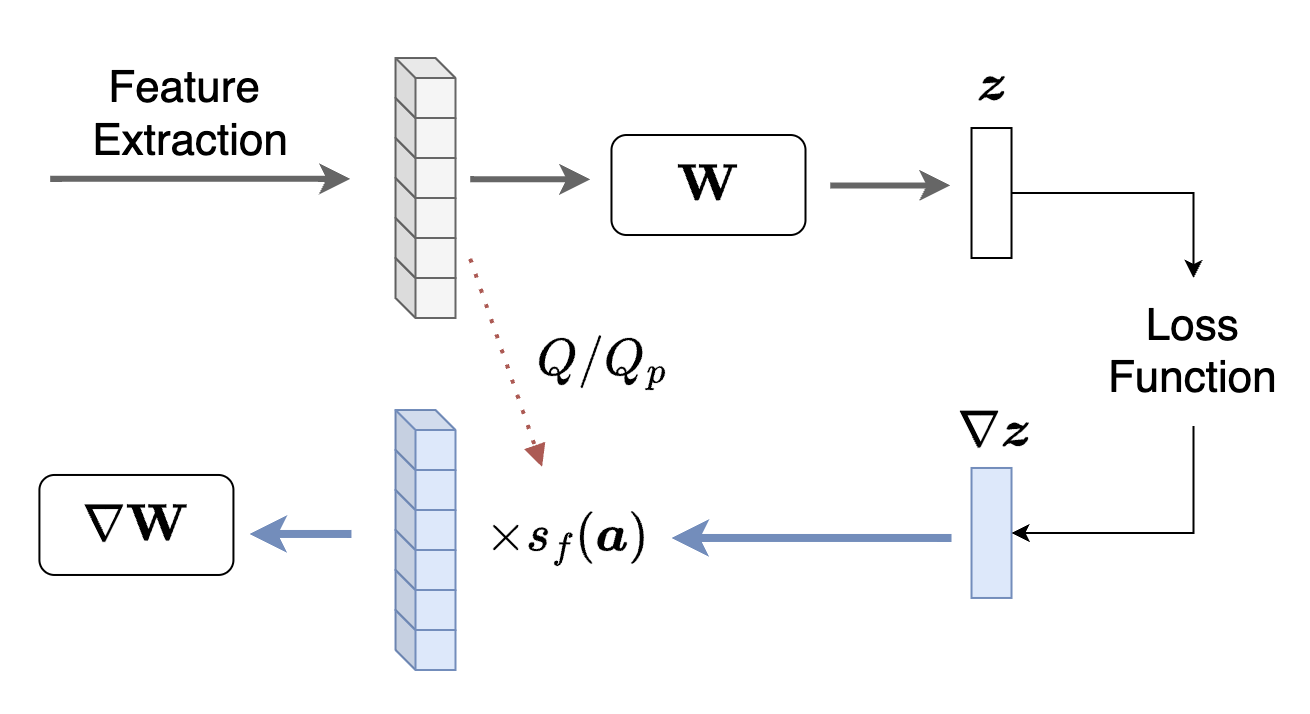}
        \caption[b]{The process of ISH training. During training, we keep the forward pass unchanged. In the backward pass, we scale activations for parameter optimization weighted by $s_f(\pF_i)$, which varies for different samples and reflects sample's ID-ness.}\label{fig:ish}
     \end{subfigure}
     \caption{Illustrations of our post-hoc model enhancement method SCALE and training time model enhancement method  ISH.}\label{fig:scale_ish}
\end{figure}

Fig.~\ref{fig:scale_ish}a illustrates how SCALE works. SCALE applies the same scaling factor $r$ as ASH, based on percentile $p$. Instead of pruning, it retains and scales \emph{all} the activations.  Doing so has two benefits. First, it enhances the separation in energy scores between ID and OOD samples. Secondly, scaling all activations equally preserve the ordinality of the logits $\logit'$ compared to $\logit$.  As such, the $\arg\max$ is not affected and there is no trade-off for ID accuracy; this is not the case with rectification, be it pruning, like in ASH or clipping, or like ReAct (see Fig. 1). Results in Tab. \ref{tab:resnet_50_openoodv1.5} and \ref{tab:main_cifar_results} verify that SCALE outperform ASH-S on all datasets and model architectures.

\subsection{Incorporating SCALE into Training}
In practice, the semantic shift of ID versus OOD data may be ambiguous.  
For example, iNaturalist dataset features different species of plants; similar objects may be found in ImageNet. 
Our hypothesis is that, during training, we can emphasize the impact of samples possessing the most distinctive in-distribution characteristics, denoted as "ID-ness". Quantifying the ID-ness of specific samples is a challenging task, so we rely on a well-trained network to assist us in this endeavor. In particular, for a well-trained network, we can reacquire the activations of all training samples. We proceed on the assumption that the normalized ID activations are greater than those of out-of-distribution (OOD) activations. To measure the degree of ID-ness within the training data, we compute their scale factor, represented as ${Q}/{Q_p}$. Armed with this measurement of ID-ness, we can then undertake the process of re-optimizing the network using the high ID-ness data. Our approach draws inspiration from the concept of intermediate tensor compression found in memory-efficient training methods~\citep{DBLP:conf/iclr/ChenXWCY23/DropIT}, where modifications are exclusively applied to the backward pass, leaving the forward pass unchanged.

Fig. \ref{fig:scale_ish}b illustrates our training time enhancement methods for OOD detection. We finetune a well-trained network, by introducing a modification to the gradient of the weights of the fully connected layer. The modified gradient is defined as follows:
\begin{equation}
 \mathbf{W}^{t+1} = \mathbf{W}^t - \eta \sum_i [(\pF_i\circ s_f(\pF_i))^{\top}\nabla{\logit_i}  ]
\end{equation} 
where $i$ denotes sample index in the batch, $\nabla$ denotes the gradient regarding to the cross entropy loss, $t$ denotes the training step $t$, and $\eta$ represents the learning rate.

Modifying activations exclusively in the backward pass offers several advantages. Firstly, it leaves the forward pass unaffected, resulting in only a minimal loss in ID accuracy. Secondly, the model architecture remains exactly the same during inference, making this training strategy compatible with any OOD post-processing techniques. Since the saved activations in the backward pass are also referred to as intermediate tensors, we term this method as Intermediate tensor SHaping (ISH).

\section{Experiments}

\subsection{Settings}\label{sec:exp_setting}

To verify SCALE as a post-hoc OOD method, we conduct experiments using 
CIFAR10, CIFAR100~\citep{Krizhevsky2009LearningML/CIFAR}, and ImageNet-1k~\citep{DBLP:conf/cvpr/DengDSLL009/ImageNet} as in-distribution (ID) data sources.

\textbf{CIFAR.} 
We used SVHN~\citep{37648/SVHN}, LSUN-Crop~\citep{DBLP:journals/corr/YuZSSX15/LSUN}, LSUN-Resize~\citep{DBLP:journals/corr/YuZSSX15/LSUN}, iSUN~\citep{DBLP:journals/corr/XuEZFKX15/iSUN}, Places365~\citep{DBLP:journals/pami/ZhouLKO018/Place}, and Textures~\citep{DBLP:conf/cvpr/CimpoiMKMV14/Textures} as OOD datasets, 
For consistency with previous work, we use the same model architecture and pretrained weights, namely, DenseNet-101~\citep{DBLP:conf/cvpr/HuangLMW17/DenseNet}, in accordance with the other post-hoc approaches DICE, ReAct, and ASH. 
Table \ref{tab:main_cifar_results} compares the 
FPR@95 and AUROC averaged across all six datasets; detailed results are provided in Appendix~\ref{app:exps}. 

\textbf{ImageNet.} In our ImageNet experiments, we follow the OpenOOD v1.5~\citep{DBLP:journals/corr/abs-2306-09301/OpenOODv1.5} benchmark, which separates OOD datasets as near-OOD and far-OOD groups. We employed SSB-hard~\citep{DBLP:conf/iclr/Vaze0VZ22/SSB} and NINCO~\citep{DBLP:conf/icml/BitterwolfM023/NINCO} as near-OOD datasets and iNaturalist~\citep{DBLP:conf/cvpr/HornASCSSAPB18/INaturalist}, Textures~\citep{DBLP:conf/cvpr/CimpoiMKMV14/Textures}, and OpenImage-O~\citep{DBLP:conf/cvpr/Wang0F022/OpenImage-O} as far-OOD datasets. Our reported metrics are the average FPR@95 and AUROC values across these categories; detailed results are given in Appendix~\ref{app:exps}. The OpenOOD benchmark includes improved hyperparameter selection with a dedicated OOD validation set to prevent overfitting to the testing set. Additionally, we provide results following the same dataset and test/validation split settings as ASH and ReAct in the appendix. We adopted the ResNet50~\citep{DBLP:conf/cvpr/HeZRS16/ResNet} model architecture and obtained the pretrained network from the torchvision library.

\textbf{Metrics.} We evaluate with two measures.  
The first is FPR@95, which measures the false positive rate at a fixed true positive rate of 95\%; lower scores are better).  The second is AUROC (Area under the ROC curve). It represents the probability that a positive in-distribution (ID) sample will have a higher detection score than a negative out-of-distribution (OOD) sample; higher scores indicate superior discrimination.

\subsection{SCALE for Post-Hoc OOD Detection}

Comparison of ODD score methods and post-hoc model enhancement methods (separated with a solid line) on the ImageNet and CIFAR are illustrated in the Table \ref{tab:resnet_50_openoodv1.5} and \ref{tab:main_cifar_results}. Notably, SCALE attains the highest OOD detection scores. 

\textbf{OOD Detection Accuracy.}
Compared to the current state-of-the-art ASH-S, SCALE demonstrates significant improvements on ImageNet -- 1.73 on Near-OOD and 0.26 on far-OOD when considering AUROC.  For FPR@95, it outperforms ASH-S by 2.27 and 0.33.
On CIFAR10 and CIFAR100, SCALE has even greater improvements of 2.48 and 2.41 for FPR@95, as well as 0.66 and 0.72 for AUROC, respectively.

\textbf{ID Accuracy.}
One of SCALE's key advantages is it only applies linear transformations on features, so ID accuracy is guaranteed to stay the same.  
This differentiates it from other post-hoc enhancement methods that rectify or prune activations, thereby modifying inference and invariably compromises the ID accuracy. 
SCALE's performance surpasses ASH-S by a substantial margin of 0.67 on the ID dataset, ImageNet-1k. This capability is pivotal for establishing a unified pipeline that excels for ID and OOD. 

\begin{table}[hbt!]
\centering
\resizebox{0.9\textwidth}{!}{
\begin{tabular}{ccccccc}
\toprule
\multirow{3}{*}{\textbf{Model}} & \multirow{3}{*}{\textbf{Postprocessor}}  & \multicolumn{2}{c}{\textbf{Near-OOD}} &\multicolumn{2}{c}{\textbf{Far-OOD}} &\multirow{2}{*}{\textbf{ID \ ACC}} \\ 

&  & FPR@95 & AUROC & FPR@95 & AUROC \\
&  & $\downarrow$ & $\uparrow$ & $\downarrow$ & $\uparrow$ & $\uparrow$ \\
\midrule
\multirow{10}{*}{ResNet50} 
& \multicolumn{1}{l}{EBO~\citep{DBLP:conf/nips/LiuWOL20/EBO}} & 68.56 & 75.89 & 38.40 & 89.47 & 76.18\\
& \multicolumn{1}{l}{MSP~\citep{DBLP:conf/iclr/HendrycksG17/MSP}}               & 65.67 & 76.02 & 51.47 & 85.23 & 76.18\\
& \multicolumn{1}{l}{MLS~\citep{DBLP:conf/icml/HendrycksBMZKMS22/MLS}}               & 67.82 & 76.46 & 38.20 & 89.58 & 76.18\\
& \multicolumn{1}{l}{GEN~\citep{DBLP:conf/cvpr/LiuLZ23/GEN}}               & 65.30 & 76.85 & 35.62 & 89.77 & 76.18 \\
& \multicolumn{1}{l}{RMDS~\citep{DBLP:journals/corr/abs-2106-09022/RMDS}}              & 65.04 & 76.99 & 40.91 & 86.38 & 76.18\\
\cmidrule{2-7}
& \multicolumn{1}{l}{TempScale~\citep{DBLP:conf/icml/GuoPSW17/TempScale}}         & 64.51 & 77.14 & 46.67 & 87.56 & 76.18 \\
& \multicolumn{1}{l}{ReAct~\citep{DBLP:conf/nips/SunGL21/ReAct}}             & 66.75 & 77.38 & 26.31 & 93.67 & 75.58 \\ 
& \multicolumn{1}{l}{ASH-S~\citep{DBLP:conf/iclr/DjurisicBAL23/ASH}}             & 62.03 & 79.63 & 16.86 & 96.47 & 75.51 \\ 

& \multicolumn{1}{l}{\textbf{SCALE (Ours)}}    & \textbf{59.76} & \textbf{81.36} & \textbf{16.53} & \textbf{96.53} & \textbf{76.18}\\ 
\bottomrule
\end{tabular}}
\caption{\textbf{OOD detection results on ImageNet-1K benchmarks.} Model choice and protocol are the same as existing works. SCALE outperforms other OOD score methods and post-hoc model enhancement methods, achieving the highest OOD detection scores and excelling in the ID-OOD trade-off. Detailed results for each dataset are given in Appendix~\ref{app:exps}.}
\label{tab:resnet_50_openoodv1.5}

\end{table}

\begin{table}[hbt!]
\centering 
\resizebox{0.8\textwidth}{!}{
\begin{tabular}{c c c c  c c}
\toprule
 & &\multicolumn{2}{c}{\textbf{CIFAR-10}} & \multicolumn{2}{c}{\textbf{CIFAR-100}} \\
\textbf{Model}&\textbf{Postprocessor} & FPR@95 & AUROC & FPR@95 & AUROC \\
& & $\downarrow$ & $\uparrow$ &  $\downarrow$ & $\uparrow$ \\
\midrule 
\multirow{6}{*}{DenseNet-101} 
&\multicolumn{1}{l}{MSP} & 48.73 & 92.46 & 80.13 & 74.36 \\
&\multicolumn{1}{l}{EBO} & 26.55 & 94.57 & 68.45 & 81.19 \\
&\multicolumn{1}{l}{ReAct} & 26.45 & 94.95 & 62.27 & 84.47 \\
&\multicolumn{1}{l}{DICE} & ${20.83}^{\pm1.58}$ & ${95.24}^{\pm0.24}$ & ${49.72}^{\pm1.69}$ & ${87.23}^{\pm0.73}$ \\
&\multicolumn{1}{l}{ASH-S}  & 15.05 & 96.61 & 41.40 & 90.02\\

&\multicolumn{1}{l}{\textbf{SCALE (Ours)}} & \textbf{12.57} & \textbf{97.27} & \textbf{38.99} & \textbf{90.74}\\
\bottomrule
\end{tabular}
}
\caption{\textbf{OOD detection results on CIFAR benchmarks.} SCALE outperform all postprocessors. Detailed results for each dataset are in the appendix.}\label{tab:main_cifar_results}
\end{table}

\textbf{Comparison with TempScale.}
Temperature scaling (TempScale) is widely used for confidence calibration~\citep{DBLP:conf/icml/GuoPSW17/TempScale}. 
SCALE and TempScale both leverage scaling for OOD detection, but with two distinctions. 
Firstly, TempScale directly scales logits for calibration, whereas SCALE applies scaling at the penultimate layer. Secondly, TempScale employs a uniform scaling factor for all samples, whereas SCALE applies a sample-specific scaling factor based on the sample's activation statistics. The sample-specific scaling is a crucial differentiator that enables the discrimination between ID and OOD samples. Notably, our SCALE model significantly outperforms TempScale in both Near-OOD and Far-OOD scenarios.

\textbf{SCALE with different percentiles $p$.}
Table \ref{tab:resnet_50_openoodv1.5} uses $p=0.85$ for SCALE and ASH-S, which is verified on the validation set. 
As detailed in Section~\ref{sec:analysis}, in order to ensure the validity of scaling, it is essential for the percentile value $p$ to fall within a specific range where the parameter $C(p)$ exhibits a sufficiently high value to meet the required condition. Our experimental observations align with this theoretical premise. Specifically, we have empirically observed that, up to the 85\% percentile threshold, the AUROC values for both Near-OOD and Far-OOD scenarios consistently show an upward trend. However, a noticeable decline becomes apparent beyond this percentile threshold. This empirical finding corroborates our theoretical insight, indicating that the parameter $C(p)$ experiences a reduction in magnitude as $p$ approaches the 90\%.

\vspace{2em}
\begin{table}[h!]
\centering 
\resizebox{\textwidth}{!}{
\begin{tabular}{c c c c  c c c c}
\toprule

$p$& 65 & 70 & 75 & 80 & 85 & 90 & 95 \\
\midrule 
Near-OOD&62.45 / 79.31 & 61.65 / 79.83 & 61.12 / 80.41  & 60.12 / 81.01 & \textbf{59.76 / 81.36} & 63.19 / 80.14 & 78.62  73.40\\
Far-OOD& 24.08 / 94.43 & 22.21 / 95.02 & 20.20 / 95.61 & 18.26 / 96.17 & \textbf{16.53 / 96.53} & 18.58 / 96.20 & 32.42  93.28\\
\bottomrule
\end{tabular}
}
\caption{FPR@95 / AUROC results on ImageNet benchmarks under different $p$.}
\end{table}

\subsection{ISH for Training-Time Model Enhancement}
We used the same dataset splits as the post-hoc experiments in Sec.~\ref{sec:exp_setting}. For training, we fine-tuned the torchvision pretrained model with ISH for 10 epochs with a cosine annealing learning rate schedule initiated at 0.003 and a minimum of 0. We additionally observed that using a smaller weight decay value (5e-6) enhances OOD detection performance. %, and thus, we selected for this purpose. 
The results are presented in Table \ref{tab:ish_openoodv1.5}. We compare ISH with other training time model enhancement methods.

{\textbf{Comparison with OOD traning methods.}} 

The work LogitNorm\citep{DBLP:conf/icml/WeiXCF0L22/LogitNorm} focuses on diagnosing the gradual narrowing of the gap between the logit magnitudes of ID and OOD distributions during later stages of training. Their proposed approach involves normalizing logits, and the scaling factor is applied within the logits space during the backward pass.

The key distinction between their LogitNorm method and our ISH approach lies in the purpose of scaling. LogitNorm scales logits primarily for confidence calibration, aiming to align the model's confidence with the reliability of its predictions. In contrast, ISH scales activations to prioritize weighted optimization, emphasizing the impact of high ID-ness data on the fine-tuning process.

\textbf{Comparisons with data augmentation-based methods.}
~\cite{DBLP:journals/corr/abs-2306-09301/OpenOODv1.5} indicates that data augmentation methods, while not originally designed for OOD detection improvement, can simultaneously enhance both ID and OOD accuracy. 

In comparison to AugMix and RegMixup, our ISH approach, while slightly reducing ID accuracy, delivers superior OOD performance with significantly fewer computational resources. When compared to AugMix, ISH achieves substantial improvements, enhancing AUROC by 0.46 and 0.8 for Near-OOD and Far-OOD, respectively, with just 0.1x the extended training epochs. Notably, ISH sets the highest AUROC records, reaching 84.01\% on Near-OOD scores and 96.79\% on Far-OOD scores among all methods on~\href{https://zjysteven.github.io/OpenOOD/}{OpenOODv1.5 benchmark}.

\begin{table}[hbt!]
\centering
\resizebox{\textwidth}{!}{
\begin{tabular}{ccccccccc}
\toprule
\multirow{3}{*}{\textbf{Model}} & \multirow{3}{*}{\textbf{Training}}& \multirow{3}{*}{\makecell{\textbf{Epochs} \\ Ori.+\textit{Ext.}}}&\multirow{3}{*}{\textbf{Postprocessor}}  & \multicolumn{2}{c}{\textbf{Near-OOD}} &\multicolumn{2}{c}{\textbf{Far-OOD}} &\multirow{2}{*}{\textbf{ID \ ACC}} \\ 

&  &  & &FPR@95 & AUROC & FPR@95 & AUROC \\
&  &  & &$\downarrow$ & $\uparrow$ & $\downarrow$ & $\uparrow$ & $\uparrow$ \\
\midrule
\multirow{6}{*}{ResNet50} 
& \multicolumn{1}{l}{LogitNorm~\citep{DBLP:conf/icml/WeiXCF0L22/LogitNorm}}   & 90+30  & MSP & 68.56 & 74.62 & 31.33 & 91.54  & 76.45\\

& \multicolumn{1}{l}{CIDER~\citep{DBLP:conf/iclr/MingSD023/CIDER}} & 90+30 & KNN & 71.69 & 68.97 & 28.69 & 92.18 & - \\

\cmidrule{2-9}
& \multicolumn{1}{l}{TorchVision Model} & 90 & SCALE & 59.76 & 81.36 & 16.53 & 96.53 & 76.13\\
& \multicolumn{1}{l}{TorchVision Model Extended} & 90+10 & SCALE & 59.25 & 82.67 & 18.48 & 96.24 & 76.84\\
& \multicolumn{1}{l}{RegMixup~\citep{DBLP:conf/nips/PintoYLTD22/RegMixup}} & 90+30 & SCALE & 63.55 & 80.85 & 19.87 & 95.94 & 76.88\\

& \multicolumn{1}{l}{AugMix~\citep{DBLP:conf/iclr/HendrycksMCZGL20/AugMix}} & 180 & SCALE & 60.58 & 83.55 & 21.01 & 95.99 & \textbf{77.64} \\

& \multicolumn{1}{l}{\textbf{ISH (Ours)}} & 90+\textbf{10} & SCALE & \textbf{55.73} & \textbf{84.01} & \textbf{15.62} & \textbf{96.79} & 76.74\\
\bottomrule
\end{tabular}}
\caption{Comparisons with data augmentation-based methods on ImageNet-1K. 
Our ISH method achieves the highest scores for both Near-OOD and Far-OOD with the shortest training epochs. "Ori." denotes the original training epochs for the pretrained network, while "Ext." denotes the extended training epochs in our training scheme.}
\label{tab:ish_openoodv1.5}

\end{table}

\section{Conclusion}

In this paper, we have conducted an in-depth investigation into the efficacy of scaling techniques in enhancing out-of-distribution (OOD) detection. Our study is grounded in the analysis of activation distribution disparities between in-distribution (ID) and OOD data. To this end, we introduce SCALE, a post-hoc model enhancement method that achieves state-of-the-art OOD accuracy when integrated with energy scores, without compromising ID accuracy. Furthermore, we extend the application of scaling to the training phase, introducing ISH, a training-time enhancement method that significantly bolsters OOD accuracy.

\bibliography{ref}
\bibliographystyle{iclr2024_conference}

\appendix

\section{Details of Proof}\label{app:proof}

\begin{manualtheorem}{3.1}
\label{prop_ratio}
Assume that ID activations $\pF_j^{(\textit{ID})} \sim \mathcal{N}^{R}(\mu^{\textit{ID}}, \sigma^{\textit{ID}})$ and OOD activations $\pF_j^{(\textit{OOD})} \sim \mathcal{N}^{R}(\mu^{\textit{OOD}}, \sigma^{\textit{OOD}})$ where $\mathcal{N}^{R}$ denotes a rectified Gaussian distribution. 
If $\mu^\text{ID}/\sigma^\text{ID}>\mu^\text{OOD}/\sigma^\text{OOD}$, then there is a range of percentiles $p$ for which a factor $ C(p)=\frac{\varphi(\sqrt{2}\operatorname {erf} ^{-1}(2p-1))}{1-\Phi(\sqrt{2}\operatorname {erf} ^{-1}(2p-1))}$ is large enough such that ${Q_p^{\text{ID}}}/{Q^{\text{ID}}}<{Q_p^{\text{OOD}}}/{Q^{\text{OOD}}}$. 
% \end{proposition}
\end{manualtheorem}

\begin{proof}
The proof schema is to derive equivalent conditions. Under the assumption that data in the latent space follows an independent and identically distributed (IID) Gaussian distribution prior to the ReLU activation (\cite{DBLP:conf/nips/SunGL21/ReAct}), we can derive that each coefficient $\pF_j^{(\textit{ID})} \sim \mathcal{N}^{R}(\mu^{\textit{ID}}, \sigma^{\textit{ID}})$ and OOD activations $\pF_j^{(\textit{OOD})} \sim \mathcal{N}^{R}(\mu^{\textit{OOD}}, \sigma^{\textit{OOD}})$ where $\mathcal{N}^{R}$ denotes a rectified Gaussian distribution.
Moreover if we denote high activation $\vh_j^{(\textit{ID})} = \pF_j^{(\textit{ID})}$ if $ \pF_j>P_{p}(\pF) $ and zeros elsewhere. Then we have $ \vh_j^{(\textit{ID})} \sim \mathcal{N}^{T}(\mu^{\textit{ID}}, \sigma^{\textit{ID}})$ and identically  $\vh_j^{(\textit{OOD})} \sim \mathcal{N}^{T}(\mu^{\textit{OOD}}, \sigma^{\textit{OOD}})$, where $\mathcal{N}^{T}$ denotes a truncated Gaussian distribution.
Then, we can calculate the expectations as follows:
\begin{align}
\mathbb{E}[\pF_j] &=  \mu \left[1- \Phi (-\frac{\mu}{\sigma}) \right] +\varphi (-\frac{\mu}{\sigma})\sigma\\
\mathbb{E}[\vh_j] &=  \mu + \frac{\varphi(m)}{1-\Phi(m)}\sigma, \ m = \frac{s-\mu}{\sigma}
\end{align}
Here, $\varphi (\cdot )$ is the probability density function of the standard normal distribution, and $\Phi (\cdot )$ is its cumulative distribution function. 

{$Q_p/Q=\frac{\sum_j\vh_j}{\sum_j\va_j}=\frac{\E[\vh_j](1-p)D}{\E[\va_j]D}$}.
Let us consider the notation $  \beta =(1-p)Q/Q_p=\frac{\mathbb{E}[\va_j] }{\mathbb{E}[\vh_j]}$. $Q_p^{\text{ID}}/Q^{\text{ID}}<Q_p^{\text{OOD}}/Q^{\text{OOD}}\iff\beta^{\text{ID}}>\beta^{\text{OOD}}$. So we focus on:
\begin{align}
     \beta = \frac{ \mu \left[1- \Phi (-\frac{\mu}{\sigma}) \right] +\varphi (-\frac{\mu}{\sigma})\sigma  }{ \mu +{\frac{\varphi(m)}{1-\Phi(m)}}\sigma}= \frac{  1- \Phi (-\frac{\mu}{\sigma})    }{ 1 +{\frac{\varphi(m)}{1-\Phi(m)}}\frac{\sigma}{\mu}} +
\frac{  \varphi (-\frac{\mu}{\sigma})\sigma  }{ \mu +\frac{\varphi(m)}{1-\Phi(m)}\sigma}
\end{align}
Let's introduce some notations for ease of analysis:
\begin{itemize}
    \item $\gamma = \frac{\mu}{\sigma}$
    \item $A = \Phi(-\gamma)$
    \item $B = \varphi(-\gamma)$
    \item $C = \frac{\varphi(m)}{1-\Phi(m)} = \frac{\varphi(\frac{s-\mu}{\sigma})}{1-\Phi(\frac{s-\mu}{\sigma})}$, 
\end{itemize}
With these definitions, we can express $\beta$ as:
\begin{align}
 \beta = \frac{  1- A    }{ 1 +C\gamma^{-1}} +
\frac{  B\sigma  }{ \mu +C\sigma}
\end{align}
We consider that 
$\gamma^{\text{ID}} \geq \gamma^{\text{OOD}}$ 
Hence, we also have:
\begin{itemize}
    \item $A^{\text{ID}} \leq A^{\text{OOD}}$
    \item $B^{\text{ID}} \leq B^{\text{OOD}}$
\end{itemize}
By definition we have that
$ s^{\text{ID}}(p) = {\mu^{\text{ID}} +\sigma^{\text{ID}} {\sqrt {2}}\operatorname {erf} ^{-1}(2p-1)}$  and
$ s^{\text{OOD}}(p) = {\mu^{\text{OOD}} +\sigma^{\text{OOD}} {\sqrt {2}}\operatorname {erf} ^{-1}(2p-1)}$
where $p$ is the proportion of data that we want to keep. So we have:
\begin{align}
C^{\text{ID}}(p) = \frac{\varphi(m^{\text{ID}})}{1-\Phi(m^{\text{ID}})} = \frac{\varphi(\frac{s^{\text{ID}}-\mu^{\text{ID}}}{\sigma^{\text{ID}}})}{1-\Phi(\frac{s^{\text{ID}}-\mu^{\text{ID}}}{\sigma^{\text{ID}}})}= \frac{\varphi(\sqrt{2}\operatorname {erf} ^{-1}(2p-1))}{1-\Phi(\sqrt{2}\operatorname {erf} ^{-1}(2p-1))}
\end{align}
Moreover, we can prove that 
$    C^{\text{OOD}}(p)  = \frac{\varphi(m^{\text{OOD}})}{1-\Phi(m^{\text{OOD}})} = \frac{\varphi(\frac{s^{\text{OOD}}-\mu^{\text{OOD}}}{\sigma^{\text{OOD}}})}{1-\Phi(\frac{s^{\text{OOD}}-\mu^{\text{OOD}}}{\sigma^{\text{OOD}}})}= \frac{\varphi(\sqrt{2}\operatorname {erf} ^{-1}(2p-1))}{1-\Phi(\sqrt{2}\operatorname {erf} ^{-1}(2p-1))} =  C^{\text{ID}}(p) $.

Now, if we consider the approximation:
\begin{align}
\mathbb{E}[ \va_j] \simeq  \mu \left[1- \Phi (-\frac{\mu}{\sigma}) \right] 
\end{align}
We assume that $\varphi\left(-\frac{\mu}{\sigma}\right)\sigma \approx 0$ since the sigma term is very small, and the second term is below one. With this approximation, we have:
\begin{align}
 \beta = \frac{  \gamma (1- A  )  }{ \gamma +C} 
\end{align}

We want to compare $\beta$ for in-distribution (ID) denoted $\beta^{\text{ID}}$ and out-of-distribution (OOD) data denoted $\beta^{\text{OOD}}$.
Moreover, we have:
\begin{align}
\beta^{\text{ID}} \geq \beta^{\text{OOD}}\iff
\frac{1- A^{\text{ID}}}{ 1 +C{\gamma^{\text{ID}}}^{-1}} \geq \frac{1- A^{\text{OOD}}}{ 1 +C{\gamma^{\text{OOD}}}^{-1}}
\iff
 \frac{1- A^{\text{ID}}}{1- A^{\text{OOD}}} \geq \frac{1 +C{\gamma^{\text{ID}}}^{-1}}{ 1 +C{\gamma^{\text{OOD}}}^{-1}}
 \end{align}
   We can use the approximation:
$\frac{1}{1 +C{\gamma^{\text{OOD}}}^{-1}} \simeq 1 -C{\gamma^{\text{OOD}}}^{-1}$ by applying a first-order Taylor expansion. Then we have:
\begin{align}
\frac{1- A^{\text{ID}}}{1- A^{\text{OOD}}} &\geq (1 +C{\gamma^{\text{ID}}}^{-1})( 1 -C{\gamma^{\text{OOD}}}^{-1})\\
&\geq 1 +C({\gamma^{\text{ID}}}^{-1} -{\gamma^{\text{OOD}}}^{-1}) - C^2({\gamma^{\text{ID}}}^{-1} {\gamma^{\text{OOD}}}^{-1})
\end{align}
    Note that by definition $C$ should be positive. 
   The given inequality can be expressed as:
   \begin{align}
   \frac{1- A^{\text{ID}}}{1- A^{\text{OOD}}}- 1 -C({\gamma^{\text{ID}}}^{-1} -{\gamma^{\text{OOD}}}^{-1}) + C^2({\gamma^{\text{ID}}}^{-1} {\gamma^{\text{OOD}}}^{-1})  \geq 0
   \end{align}
   We can rewrite it as:
   \begin{align}\mathbb{a}_1C^2 + \mathbb{a}_2C + \mathbb{a}_3  \geq 0
   \end{align}
   Here we have the following notations: $\mathbb{a}_1 =({\gamma^{\text{ID}}}^{-1} {\gamma^{\text{OOD}}}^{-1})$ and $\mathbb{a}_2 =-({\gamma^{\text{ID}}}^{-1} -{\gamma^{\text{OOD}}}^{-1}) $
   and $\mathbb{a}_3= \frac{1- A^{\text{ID}}}{1- A^{\text{OOD}}}- 1$.
   Let us define $\Delta = \mathbb{a}_2^2-4\mathbb{a}_1\mathbb{a}_3$  
   Then we have:
   \begin{align}
    \Delta &= ({\gamma^{\text{ID}}}^{-1} -{\gamma^{\text{OOD}}}^{-1})^2 - 4({\gamma^{\text{ID}}}^{-1} {\gamma^{\text{OOD}}}^{-1})\left(\frac{1- A^{\text{ID}}}{1- A^{\text{OOD}}}- 1\right)\\
    &= {\gamma^{\text{ID}}}^{-2} +{\gamma^{\text{OOD}}}^{-2} - 2({\gamma^{\text{ID}}}^{-1} {\gamma^{\text{OOD}}}^{-1})\left(2\frac{1- A^{\text{ID}}}{1- A^{\text{OOD}}}- 1\right)\\
    &= \left( {\gamma^{\text{ID}}}^{-1} +{\gamma^{\text{OOD}}}^{-1} \right)^2 - 4({\gamma^{\text{ID}}}^{-1} {\gamma^{\text{OOD}}}^{-1})\left(\frac{1- A^{\text{ID}}}{1- A^{\text{OOD}}}\right)
    \end{align}
    Since $\mathbb{a}_1 > 0$, there are two possible cases:
    \begin{itemize}
        \item if $\Delta \leq 0$ then $C(p) \in \mathbb{R}^+$
        \item  if $\Delta > 0$ then $C(p) \in \left[ \max\left( \frac{({\gamma^{\text{ID}}}^{-1} -{\gamma^{\text{OOD}}}^{-1}) + \sqrt{\Delta}}{2({\gamma^{\text{ID}}}^{-1} {\gamma^{\text{OOD}}}^{-1})} , 0^+ \right),+\infty\right) $. Note that another side $({\gamma^{\text{ID}}}^{-1} -{\gamma^{\text{OOD}}}^{-1}) \leq 0$ so $({\gamma^{\text{ID}}}^{-1} -{\gamma^{\text{OOD}}}^{-1}) - \sqrt{\Delta} \leq 0$. So we do not consider this.
    \end{itemize}

    In summary, there is a valid range of pruning $p$ value satisfying the valid range of $C(p)$ so that the statistics $Q_p/Q$ of the ID distribution is smaller than that of the OOD distributions. $p$ with a larger $C(p)$ is more applicable to any case.
\end{proof}

\section{Full Experiments}\label{app:exps}
In this section, we provide full results for SCALE post-hoc model enhancement. Tab. \ref{tab:full_openood_imagenet} shows full results on ImageNet and Tab. \ref{tab:detailresultscifar10} and \ref{tab:detailresultscifar100} show full results on CIFAR10 and CIFAR100. We also provide ImageNet results following dataset setting of ReAct and ASH in Tab. \ref{tab:ash_datasets} for more comparison.

\begin{table}[h!]
\centering
\resizebox{\textwidth}{!}{\begin{tabular}{@{}c|ccc|cccc|c@{}}
\toprule
\multirow{2}{*}{ResNet50 }   & \multicolumn{3}{c|}{Near-OOD}                                         & \multicolumn{4}{c|}{Far-OOD}                                                              & \multirow{2}{*}{ID \ Accuracy} \\ 
                                        &  SSB-hard            & NINCO                  & Average               & iNaturalist           & Textures              & OpenImage-O               &  Average                                      \\
\midrule 
EBO       & 76.54 / 72.08 & 60.59 / 79.70 & 68.56 / 75.89 & 31.33 / 90.63 & 45.77 / 88.7 & 38.08 / 89.06 & 38.40 / 89.47 & 76.18                   \\ 
MSP       & 74.49 / 72.09 & 56.84 / 79.95 & 65.67 / 76.02 & 43.34 / 88.41 & 60.89 / 82.43 & 50.16 / 84.86 & 51.47 / 85.23 & 76.18                   \\ 
MLS       & 76.19 / 72.51 & 59.49 / 80.41 & 67.84 / 76.46 & 30.63 / 91.16 & 46.11 / 88.39 & 37.86 / 89.17 & 38.20 / 89.58 & 76.18                   \\
GEN       & 75.72 / 72.01 & 54.88 / 81.70 & 65.30 / 76.85 & 26.12 / 92.44 & 46.23 / 87.60 & 34.52 / 89.26 & 35.62 / 89.77 & 76.18                   \\
RMDS       &  77.88 / 71.77 &  52.20 / 82.22 &  65.04 / 76.99 &  33.67 / 87.24 &  48.80 / 86.08 &  40.27 / 85.84 &  40.91 / 86.38 & 76.18                   \\ 
\midrule
TempScale       & 73.90 / 72.87 & 55.12 / 81.41 & 64.51 / 77.14 & 37.70 / 90.50 & 56.92 / 84.95 & 45.39 / 87.22 & 46.67 / 87.56 & 76.18                   \\ 
ReAct       & 77.57 / 73.02 & 55.92 / 81.73 & 66.75 / 77.38 & 16.73 / 96.34 & 29.63 / 92.79 & 32.58 / 91.87 & 26.31 / 93.67 & 75.58                   \\ 
ASH-S       & 70.80 / 74.72 & 53.26 / 84.54 & 62.03 / 79.63 & 11.02 / 97.72 & \textbf{10.90} / \textbf{97.87} & 28.60 / 93.82 &  16.86 / 96.47 & 75.51                   \\ 
SCALE (Ours)       & \textbf{67.72} / \textbf{77.35} & \textbf{51.80} / \textbf{85.37} & \textbf{59.76} / \textbf{81.36} & \textbf{9.51} / \textbf{98.02} & 11.90 / 97.63 & \textbf{28.18} / \textbf{93.95} & \textbf{16.53} / \textbf{96.53} & \textbf{76.18}                   \\ 
\bottomrule
\end{tabular}}
\caption{FPR95 $\downarrow$ / AUROC $\uparrow$ for ResNet50 on ImageNet on OpenOOD v1.5 benchmark. }\label{tab:full_openood_imagenet}
\end{table}

\begin{table}[hbt!]
\resizebox{\textwidth}{!}{\begin{tabular}{c c c c c c c c c c c c c}
\toprule
 & & \multicolumn{8}{c}{\textbf{OOD Datasets}} & & \\
\cmidrule{3-10} \\
\textbf{Model} & \textbf{Methods} & \multicolumn{2}{c}{\textbf{iNaturalist}} & \multicolumn{2}{c}{\textbf{SUN}} & \multicolumn{2}{c}{\textbf{Places}} & \multicolumn{2}{c}{\textbf{Textures}} & \multicolumn{2}{c}{\textbf{Average}} & \textbf{ID ACC} \\
& & & & & & & & & & & \\
& & FPR95 $\downarrow$ & AUROC $\uparrow$ & FPR95 $\downarrow$ & AUROC $\uparrow$ & FPR95 $\downarrow$ & AUROC $\uparrow$ & FPR95 $\downarrow$ & AUROC $\uparrow$ & FPR95 $\downarrow$ & AUROC $\uparrow$ \\
\midrule

\multirow{7}{*}{ResNet50} & \multicolumn{1}{l}{MSP} & 54.99 & 87.74 & 70.83 & 80.86 & 73.99 & 79.76 & 68.00 & 79.61 & 66.95 & 81.99 & 76.12\\
& \multicolumn{1}{l}{EBO} & 55.72 & 89.95 & 59.26 & 85.89 & 64.92 & 82.86 & 53.72 & 85.99 & 58.41 & 86.17 & 76.12\\
& \multicolumn{1}{l}{ReAct} & 20.38 & 96.22 & 24.20 & 94.20 & \textbf{33.85} & 91.58 & 47.30 & 89.80 & 31.43 & 92.95 & -\\
& \multicolumn{1}{l}{DICE} & 25.63 & 94.49 & 35.15 & 90.83 & 46.49 &  87.48 & 31.72 & 90.30 & 34.75 & 90.77 & -\\
& \multicolumn{1}{l}{DICE + ReAct} & 18.64 & 96.24 & 25.45 & 93.94 & 36.86 & 90.67 & 28.07 & 92.74 & 27.25 & 93.40 & -\\
& \multicolumn{1}{l}{ASH-S} & 11.49 & 97.87 & 27.98 & 94.02 & 39.78 & 90.98 & \textbf{11.93} & \textbf{97.60} & 22.80 & \textbf{95.12} & 74.98 \\
\cmidrule{2-13}
& \multicolumn{1}{l}{\textbf{SCALE (Ours)} } & \textbf{9.50} & \textbf{98.17} & \textbf{23.27} & \textbf{95.02} & 34.51 & \textbf{92.26} & 12.93 & 97.37 & \textbf{20.05} & \textbf{95.71} & \textbf{76.12}\\
\bottomrule

\end{tabular}}
\caption{OOD detection results for ResNet 50 following the exact same metrics and testing splits as~\cite{DBLP:conf/nips/SunGL21/ReAct}. ResNet is trained with ID data (ImageNet-1k) only. $\uparrow$ indicates larger values are better and $\downarrow$ indicates smaller values are better. All values are percentages. SCALE consistently perform better than ASH-S, across all the OOD datasets. }\label{tab:ash_datasets}

\end{table}

\begin{sidewaystable}
\centering
\caption{\footnotesize Detailed results for \textbf{CIFAR-10}.}
\scalebox{0.75}{
\begin{tabular}{lllllllllllllllc} \toprule
\multirow{2}{*}{\textbf{Method}} & \multicolumn{2}{c}{\textbf{SVHN}} & \multicolumn{2}{c}{\textbf{LSUN-c}} & \multicolumn{2}{c}{\textbf{LSUN-r}} & \multicolumn{2}{c}{\textbf{iSUN}} & \multicolumn{2}{c}{\textbf{Textures}} & \multicolumn{2}{c}{\textbf{Places365}} & \multicolumn{2}{c}{\textbf{Average}} & \multirow{2}{*}{\textbf{ID ACC}} \\ \cmidrule{2-15}
 & \textbf{FPR95} & \textbf{AUROC} & \textbf{FPR95} & \textbf{AUROC} & \textbf{FPR95} & \textbf{AUROC} & \textbf{FPR95} & \textbf{AUROC} & \textbf{FPR95} & \textbf{AUROC} & \textbf{FPR95} & \textbf{AUROC} & \textbf{FPR95} & \textbf{AUROC}  \\ 
 & \quad $\downarrow$ & \quad $\uparrow$ & \quad $\downarrow$ & \quad $\uparrow$ & \quad  $\downarrow$ & \quad $\uparrow$ & \quad $\downarrow$ & \quad  $\uparrow$ &  \quad $\downarrow$ & \quad $\uparrow$ & \quad $\downarrow$ & \quad $\uparrow$ & \quad $\downarrow$ & \quad $\uparrow$ & \quad $\uparrow$ \\ \midrule
MSP  & 47.24 & 93.48 & 33.57 & 95.54 & 42.10 & 94.51 & 42.31 & 94.52 & 64.15 & 88.15 & 63.02 & 88.57 & 48.73 & 92.46 & 94.53\\
EBO & 40.61 & 93.99 & 3.81 & 99.15 & 9.28 & 98.12 & 10.07 & 98.07 & 56.12 & 86.43 & 39.40 & 91.64 & 26.55 & 94.57 & 94.53\\ 
ReAct  & 41.64 & 93.87 & 5.96 & 98.84 & 11.46 & 97.87 & 12.72 & 97.72 & 43.58 & 92.47 & 43.31 & 91.03 & 26.45 & 94.67 & -\\
 {DICE} 
 & 25.99$^{\pm{5.10}}$ & 95.90$^{\pm{1.08}}$ & 0.26$^{\pm{0.11}}$ & 99.92$^{\pm{0.02}}$ & 3.91$^{\pm{0.56}}$ & 99.20$^{\pm{0.15}}$ & 4.36$^{\pm{0.71}}$ & 99.14$^{\pm{0.15}}$ & 41.90$^{\pm{4.41}}$ & 88.18$^{\pm{1.80}}$ & 48.59$^{\pm{1.53}}$ & 89.13$^{\pm{0.31}}$ & 20.83$^{\pm{1.58}}$ & 95.24$^{\pm{0.24}}$ & -\\
 
ASH-S  & 6.51 & 98.65 & 0.90 & 99.73 & 4.96 & 98.92 & 5.17 & 98.90 & 24.34 & 95.09 & 48.45 & 88.34 & 15.05 & 96.61 & 94.02 \\
\midrule
 
SCALE (Ours)  & 5.80 & 98.72 & 0.73 & 99.74 & 3.36 & 99.22 & 3.43 & 99.21 & 23.42 & 94.97 & 38.69 & 91.74 & 12.57 & 97.27 & 94.53 \\
\bottomrule

\end{tabular}}
\label{tab:detailresultscifar10}
\vspace{2\baselineskip}
\caption{Detailed results for \textbf{CIFAR-100}. }
\scalebox{0.75}{
\begin{tabular}{lllllllllllllllc} \toprule
\multirow{2}{*}{\textbf{Method}} & \multicolumn{2}{c}{\textbf{SVHN}} & \multicolumn{2}{c}{\textbf{LSUN-c}} & \multicolumn{2}{c}{\textbf{LSUN-r}} & \multicolumn{2}{c}{\textbf{iSUN}} & \multicolumn{2}{c}{\textbf{Textures}} & \multicolumn{2}{c}{\textbf{Places365}} & \multicolumn{2}{c}{\textbf{Average}} & \multirow{2}{*}{\textbf{ID ACC}} \\ \cmidrule{2-15}
 & \textbf{FPR95} & \textbf{AUROC} & \textbf{FPR95} & \textbf{AUROC} & \textbf{FPR95} & \textbf{AUROC} & \textbf{FPR95} & \textbf{AUROC} & \textbf{FPR95} & \textbf{AUROC} & \textbf{FPR95} & \textbf{AUROC} & \textbf{FPR95} & \textbf{AUROC}  \\ 
 & \quad $\downarrow$ & \quad $\uparrow$ & \quad $\downarrow$ & \quad $\uparrow$ & \quad  $\downarrow$ & \quad $\uparrow$ & \quad $\downarrow$ & \quad  $\uparrow$ &  \quad $\downarrow$ & \quad $\uparrow$ & \quad $\downarrow$ & \quad $\uparrow$ & \quad $\downarrow$ & \quad $\uparrow$ & \quad $\uparrow$ \\ \midrule
MSP & 81.70 & 75.40 & 60.49 & 85.60 & 85.24 & 69.18 & 85.99 & 70.17 & 84.79 & 71.48 & 82.55 & 74.31 & 80.13 & 74.36 & 75.04\\
EBO & 87.46 & 81.85 & 14.72 & 97.43 & 70.65 & 80.14 & 74.54 & 78.95 & 84.15 & 71.03 & 79.20 & 77.72 & 68.45 & 81.19 & 75.04\\ 
ReAct & 83.81 & 81.41 & 25.55 & 94.92 & 60.08 & 87.88 & 65.27 & 86.55 & 77.78 & 78.95 & 82.65 & 74.04 & 62.27 & 84.47 & -\\
DICE & 54.65$^{\pm{4.94}}$ & 88.84$^{\pm{0.39}}$ & 0.93$^{\pm{0.07}}$ & 99.74$^{\pm{0.01}}$ & 49.40$^{\pm{1.99}}$ & 91.04$^{\pm{1.49}}$ & 48.72$^{\pm{1.55}}$ & 90.08$^{\pm{1.36}}$ & 65.04$^{\pm{0.66}}$ & 76.42$^{\pm{0.35}}$ & 79.58$^{\pm{2.34}}$ & 77.26$^{\pm{1.08}}$ & 49.72$^{\pm{1.69}}$ & 87.23$^{\pm{0.73}}$ & -\\

ASH-S & 25.02 & 95.76 & 5.52 & 98.94 & 51.33 & 90.12 & 46.67 & 91.30 & 34.02 & 92.35 & 85.86 & 71.62 & 41.40 & 90.02  & 71.65 \\
\midrule
SCALE (Ours) & 22.05 & 96.29 & 4.48 & 99.16 & 46.02 & 91.54 & 42.14 & 92.47 & 34.20 & 92.34 & 85.04 & 72.66 & 38.99 & 90.74 & 75.04\\
\bottomrule
\end{tabular}}
\label{tab:detailresultscifar100}
\end{sidewaystable}
\end{document}

%% file: math_commands.tex
\usepackage{amsmath,amsfonts,bm}

\def\eqref#1{equation~\ref{#1}}
\def\1{\bm{1}}

\def\va{{\bm{a}}}
\def\vb{{\bm{b}}}

\def\vh{{\bm{h}}}

\def\vx{{\bm{x}}}

\def\vz{{\bm{z}}}

\DeclareMathAlphabet{\mathsfit}{\encodingdefault}{\sfdefault}{m}{sl}
\SetMathAlphabet{\mathsfit}{bold}{\encodingdefault}{\sfdefault}{bx}{n}

\newcommand{\E}{\mathbb{E}}

\DeclareMathOperator*{\argmax}{arg\,max}